# A New Open-Access Platform for Measuring and Sharing mTBI Data

**August G. Domel[1,*], Samuel J. Raymond[1,*], Chiara Giordano[1,*], Yuzhe Liu[1], Seyed Abdolmajid Yousefsani[1], Michael Fanton[2], Ileana Pirozzi[1], Ali Kight[1], Brett Avery[3], Athanasia Boumis[3], Tyler Fetters[4], Simran Jandu[3], William M Mehring[3], Sam Monga[5], Nicole Mouchawar[6], India Rangel[3], Eli Rice[3], Pritha Roy[3], Sohrab Sami[3], Heer Singh[3], Lyndia Wu[1,7], Calvin Kuo[2,8], Michael Zeineh[6], Gerald Grant[9,10], David B. Camarillo[1,2,10,⊺]**

[1] Department of Bioengineering, Stanford University, Stanford, CA, 94305, USA.
[2] Department of Mechanical Engineering, Stanford University, Stanford, CA, 94305, USA.
[3] Stanford Center for Clinical Research, Stanford University, Stanford, CA, 94305, USA.
[4] Intel Sports, Intel Corporation, Santa Clara, CA, 95054, USA.
[5] Asterix Inc., Toronto, Ontario, M4B 1B3, Canada.
[6] Department of Radiology, Stanford University, Stanford, CA, 94305, USA.
[7] Department of Mechanical Engineering, University of British Columbia, Vancouver, Canada.
[8] Department of Biomedical Engineering, University of British Columbia, Vancouver, Canada.
[9] Department of Neurology, Stanford University, Stanford, CA, 94305, USA.
[10] Department of Neurosurgery, Stanford University, Stanford, CA, 94305, USA.
[*] Co-first authors
[⊺] Corresponding author (e-mail: dcamarillo@stanford.edu).

**Authors' Contact Information**:

August G. Domel [1]
    Postal Address: 443 Via Ortega, Room 202, 94305 Stanford, CA, USA
    Email Address: agdomel@stanford.edu

Samuel J. Raymond [1]
    Postal Address: 443 Via Ortega, Room 202, 94305 Stanford, CA, USA
    Email Address: sjray@stanford.edu

Chiara Giordano [1]
    Postal Address: 443 Via Ortega, Room 202, 94305 Stanford, CA, USA
    Email Address: kiara.gio88@gmail.com

Yuzhe Liu [1]
    Postal Address: 443 Via Ortega, Room 202, 94305 Stanford, CA, USA
    Email Address: yuzheliu@stanford.edu

S. A. Yousefsani [1]
    Postal Address: 443 Via Ortega, Room 202, 94305 Stanford, CA, USA

Email Address: sam.ysani.mech@gmail.com

Michael Fanton [2]
Postal Address: 443 Via Ortega, Room 202, 94305 Stanford, CA, USA
Email Address: mfanton@stanford.edu

Ileana Pirozzi [1],
Postal Address: 443 Via Ortega, Room 202, 94305 Stanford, CA, USA
Email Address: ipirozzi@stanford.edu

Ali Kight [1]
Postal Address: 443 Via Ortega, Room 202, 94305 Stanford, CA, USA
Email Address: akight@stanford.edu

Brett Avery [3]
Postal Address: 1701 Page Mill Road, 94305 Stanford, CA, USA
Email Address: bavery@stanford.edu

Athanasia Boumis [3]
Postal Address: 1701 Page Mill Road, 94305 Stanford, CA, USA
Email Address: boumis@stanford.edu

Tyler Fetters [4]
Postal Address: 3226 N 81st Pl., 85251 Scottsdale, AZ, USA
Email Address: fetterstyler@gmail.com

Simran Jandu [3]
Postal Address: 1701 Page Mill Road, 94305 Stanford, CA, USA
Email Address:  simranjandu8@gmail.com

Will Mehring  [3]
Postal Address: 1701 Page Mill Road, 94305 Stanford, CA, USA
Email Address: wmehring@usc.edu

Sam Monga [5]
Postal Address: 10115 Scenic Boulevard, 95014 Cupertino, CA, USA
Email Address: samvitmonga@gmail.com

Nicole Mouchawar [6]
Postal Address: 300 Pasteur Dr, MC 5105, Stanford, CA, USA
Email Address: nmouchaw@stanford.edu

India Rangel [3]
Postal Address: 1701 Page Mill Road, 94305 Stanford, CA, USA

Email Address: icrangel@ucdavis.edu

Eli Rice [3]
Postal Address: 1701 Page Mill Road, 94305 Stanford, CA, USA
Email Address: riceen15@stanford.edu

Sohrab Sami [3]
Postal Address: 1701 Page Mill Road, 94305 Stanford, CA, USA
Email Address: ssami@stanford.edu

Pritha Roy [3]
Postal Address: 1701 Page Mill Road, 94305 Stanford, CA, USA
Email Address: pgroy@ucdavis.edu

Heer Singh [3]
Postal Address: 1701 Page Mill Road, 94305 Stanford, CA, USA
Email Address: heer.singh@health.slu.edu

Lyndia Wu [1,7]
Postal Address: 6250 Applied Science Lane, Rm 2059, Vancouver, BC, V6T 1Z4, Canada
Email Address: lwu@mech.ubc.ca

Calvin Kuo [2,8]
Postal Address: 2259 Lower Mall, Lower Mall Research Station, Room 350, Vancouver, BC  V6T1Z4
Email Address: calvin.kuo@ubc.ca

Michael Zeineh [6]
Postal Address: 300 Pasteur Dr, MC 5105, Stanford, CA, USA
Email Address: mzeineh@stanford.edu

Gerald Grant [9,10]
Postal Address: 300 Pasteur Dr, Room R211, MC 5327, Stanford, CA, USA
Email Address: ggrant2@stanford.edu

David B. Camarillo [1,2,10]
Postal Address: 443 Via Ortega, Room 202, 94305 Stanford, CA, USA
Email Address: dcamarillo@stanford.edu
Telephone number: (650) 725-2590

## Abstract

Despite numerous research efforts, the precise mechanisms of concussion have yet to be fully uncovered. Clinical studies on high-risk populations, such as contact sports athletes, have become more common and give insight on the link between impact severity and brain injury risk through the use of wearable sensors and neurological testing. However, as the number of institutions operating these studies grows, there is a growing need for a platform to share these data to facilitate our understanding of concussion mechanisms and aid in the development of suitable diagnostic tools. To that end, this paper puts forth two contributions: 1) a centralized, open-source platform for storing and sharing head impact data, in collaboration with the Federal Interagency Traumatic Brain Injury Research informatics system (FITBIR), and 2) a deep learning impact detection algorithm (MiGNet) to differentiate between true head impacts and false positives for the previously biomechanically validated instrumented mouthguard sensor (MiG2.0), all of which easily interfaces with FITBIR. We report 96% accuracy using MiGNet, based on a neural network model, improving on previous work based on Support Vector Machines achieving 91% accuracy, on an out of sample dataset of high school and collegiate football head impacts. The integrated MiG2.0 and FITBIR system serve as a collaborative research tool to be disseminated across multiple institutions towards creating a standardized dataset for furthering the knowledge of concussion biomechanics.

# 1. Introduction

In this study, we seek to address two major outstanding problems that we believe exist with pinning down the underlying biomechanical mechanisms of brain injury, and more specifically concussion. Because of the rarity of measuring biomechanical data during a concussion, we believe the first of two major issues is the lack of a large, high-quality head impact dataset with associated clinical outcomes. Secondly, because many of the systems, for example the Head Impact Telemetry System (HITS), used to capture the biomechanical data of head impacts often generate many false impacts, this can further generate confusing and conflicting results among studies. The only way around this is to video verify and confirm every real and fake impact measured, which is a tedious process that hampers and slows research significantly, or as is often the case, to publish the data as if all impacts are real. This is something that has hampered previous efforts at a large head impact dataset, such as that of the Concussion Assessment, Research and Education (CARE) Consortium. Given these two major issues, the goal of this study is two-fold: 1) to put forth a much needed centralized and open-source platform for storing and sharing in vivo head impact data and 2) to put forth a deep learning impact detection algorithm for differentiating between true and fake impacts of the extensively validated (i.e. high-quality) instrumented mouthguard sensor which has been designed to easily interface with the open-source platform. We seek to provide the research community with this integrated platform for collecting, processing, and sharing high-quality, standardized, biomechanical head impact data, to create a comprehensive injury and non-injury dataset. The proposed research platform is developed in collaboration with and hosted by the National Institutes of Health's (NIH) Federal Interagency Traumatic Brain Injury Research (FITBIR) informatics system [1]. Our instrumented mouthguard device collects kinematic data according to the recommended Common Data Elements (CDEs) from the National Institute of Neurological Disorders and Stroke (NINDS), which allows integration and sharing of data collected among different investigators via the platform [2]. Furthermore, our NIH R24 grant allows for the funding of 20 instrumented mouthguards for each site that joins the study to contribute to this platform.

An open-source platform, such as this, is of such great importance because of the known dangers associated with brain injury. Traumatic Brain Injury (TBI) is a leading cause of death and disability; however, more recently it has been proposed that even mild TBI (mTBI), or concussion, can increase the risk for neurodegenerative diseases, such as Alzheimer's [3, 4]. Researchers have accumulated evidence suggesting that even subconcussive impacts, previously thought to be innocuous, result in lasting neurophysiological changes [5, 6, 7]. Therefore, there is a pressing need to uncover the link between the number and severity of head impacts sustained in a lifespan, and how this correlates to acute symptomatology and affects risk of long-term chronic neurodegeneration. The scientific community would greatly benefit from a metric for concussion diagnosis, such as a potential concussion threshold. However, in order to do this, accurate biomechanical measurements of head impacts is crucial.

Previous efforts to measure head impact kinematics have leveraged sensor devices mounted to athlete's heads through headbands, ear plugs, helmets, bite bars, skin patches, and mouthguards [8-12]. However, the lack of robust coupling to the skull of some of these devices has produced inaccuracies in the measurements observed both in laboratory and field settings [11, 13-16]. Thus, it is clear that when using them with athletes in field studies, the conclusions drawn have limitations [12]. When this issue is coupled with the lack of impact detection algorithms or video verification of impacts used by many studies, many fake impacts will confuse the conclusions, and the results compared between studies can be conflicting [17-20]. Some studies have found no correlation between biomechanical metrics and the number of incurred events, severity, or time to symptoms [20, 21]. However, others report injury thresholds based on head accelerations [22, 23], which are ultimately used to rate how safe certain protective systems are that are on the market. Thus, ongoing research aimed at establishing the relationship between head impact biomechanics and brain health is hampered by the lack of a large, high-quality, reliable concussive and subconcussive dataset measured in 6 degrees of freedom [20-23].

In recent years, it has become clear that mouth wearables (i.e. instrumented mouthguards) have the tightest coupling to the skull of current head kinematic wearables [9]. Specifically, the Stanford instrumented mouthguard (herein referred to as the "MiG2.0") has been very recently extensively validated to ensure high-quality data is collected for head impacts [24]. Using this instrumented mouthguard, along with a newly developed deep learning algorithm to remove fake impacts, we believe we can produce high quality data to lead to more meaningful study conclusions. Recently, a number of studies have employed machine learning approaches to remove fake impacts [23 - 27]. Wu et al. used a support vector machine (SVM) to detect head impacts in american football players [27]. Others, such as the work conducted in [28] by Motiwale et al. have used deep learning to analyze biomechanical signals. A benefit of using deep learning tools, like neural networks, is that the features of interest need not be chosen by the researchers but instead are borne out of the data during the training process. Adding more data to the system typically improves the model more and more for these kinds of algorithms. This makes the expansion of the database ever more useful. An interesting comparison of the predictive power between SVM and deep neural networks is the choice of features. In the SVM these need to be engineered, known apriori, so that the model can use them to learn the data. However in neural networks, these features are discovered during the training process itself. The benefit of this is that the features are typically optimally chosen to explain the data, the downside being that the knowledge of what these features relate to can be difficult, if not impossible, to explain. This tradeoff between accuracy and explainaility has not yet been explored in the detection of head impacts, as it relates to mTBI, and is focussed on in this work. While the accuracy of detecting head impacts is certainly the most important metric, understanding what is sacrificed to achieve this is also valuable in terms of model selection. In this study, we present a deep learning model to analyze and remove fake impacts and compare it to an SVM to understand its benefits.

With the extensively validated MiG2.0 and its newly improved deep learning impact detection algorithms discussed in this paper, we hope for this manuscript to be a call to collaborators all over to join in the study and contribute to the new open-source platform. We hope that this platform and associated high-quality data collection will more quickly and robustly shed light on the mechanisms of brain injury and concussion.

## 2. Materials and Methods

### 2.1 The MiG. 2.0

We have developed an instrumented mouthguard, the MiG2.0, for measuring linear and angular head kinematics during an impact. Several images of the molded mouthguard as well as the electronics embedded within can be found in Fig. 2. The device is capable of sensing six degrees of freedom (6DOF) kinematics via a triaxial accelerometer (H3LIS331DL, ST Microelectronics, Geneva, Switzerland) with maximum range of ±400 g, and a triaxial gyroscope (ITG-3701, InvenSense Inc., San Jose, CA, US) with maximum range of ±4000 degrees/sec. Linear acceleration is acquired at 1000 Hz, while angular velocity is acquired at 8000Hz [17]. Data is temporarily stored in an on-board buffer, which allows the acquisition of pre- and post-trigger data associated with an event. When the linear acceleration threshold value set by the user is crossed at any of the linear acceleration components, an event is written to the on-board flash memory chip. Trigger threshold, pre-trigger time, event duration, and sample rate of the inertial measurement units (IMUs) are user adjustable. The device is also capable of detecting when it is being worn when using a proximity sensing via an infrared emitter-receiver IR pair (TMD2771, Texas Advanced Optoelectronic Solutions Inc., Plano, TX, US) which detects light reflected off of the teeth. The sensor board is completely sealed between three layers of ethylene vinyl acetate (EVA) material and communication occurs via bluetooth. A tight fit to the dentition is achieved by forming the EVA material around a player-specific dental model. IMUs are located at the incisors to isolate sensors from external disturbances. Furthermore, unique shock absorbers are placed at the molars to reduce mandible disturbances occurring in chewing, biting, and talking with the device on the teeth, as discussed by Kuo et al. [15]. It is important to note that the accuracy of all kinematic data for the MiG2.0 has been previously validated and reported in a recent manuscript [24].

### 2.2 Data download using BiteMac and upload to shared FITBIR platform

To access the kinematic data collected by the mouthguard, we developed an application, BiteMac, to interface with the MiG2.0. BiteMac is a free iOS and macOS application that enables quick, wireless, and intuitive download of kinematic data in comma separated value (CSV) format using the recommended common data elements (CDE) from the National Institute of Neurological Disorders and Strokes (NINDS) for collecting sensor data. CDEs were developed by a group of experts to facilitate collaboration across investigators, as well as interconnectivity with other informatics platforms [29]. We have worked with the Federal Interagency Traumatic Brain

Injury Research (FITBIR) agency to create an open-access platform on their website to share standardized concussion data collected using the MiG2.0 or any other head impact kinematic sensor (Methods 2.3).

Finally, we have developed a website which will enable researchers to register to use the MiG2.0 for their studies [2]. This website also contains the necessary scripts to process and standardize data output from the MiG2.0, as well as perform impact classification using our machine learning algorithm described later.

## 2.3 The FITBIR online platform

The FITBIR informatics system [30] was developed to share data (whether taken with the MiG2.0 or another similar high-quality device) across the TBI research field, in order to facilitate collaboration between laboratories and enhance interconnectivity with other informatics platforms thus allowing for aggregation, integration, and rigorous comparison of data [27]. Our integrated system provides the common data definitions and standards that are required for community-wide sharing, and qualified researchers can join the study and request access privileges to the data stored in FITBIR [30]. The FITBIR data dictionary incorporates and extends the CDE definitions developed by the NINDS CDE Workgroup, whereby domestic and international neuroscience experts developed disease-specific CDEs [31,32]. To date, CDEs have been developed for head kinematics sensors, a full list of which can be found at [1], the SCAT-5 (Sport Concussion Assessment Tool), and the ImPACT exam (Immediate Post-Concussion Assessment Testing) [1,32]. For example, these CDEs include information on the participants as well as their impact data, ranging from the participants' weight and heights to the data measured by the instrumented mouthguards and via the sideline neurocognitive tests. Data can be uploaded to our study in the FITBIR platform in CSV format, and we provide collaborating investigators with scripts to harmonize the MiG2.0 data with the FITBIR data input format. Data upload can be performed on the FITBIR website (the overall workflow for this is illustrated in Fig. 1) and should be carried out at the end of data collection by all participating investigators, once per year. Note that data is anonymized to care for all protected health information (PHI). As previously discussed in section 2.2, researchers are able to register to join the study via our website [2], and then contribute to the data repository on FITBIR upon conducting their study.

## 2.4 MiG2.0 Deployment to Generate Impact Detection Training and Testing Dataset

We deployed the MiG2.0 during practices and games to one college football team, Stanford University, and three local Bay Area high schools. We video analyzed the data from the collegiate players (n=12), including 5 outside linebackers, 1 inside linebacker, 1 running back, 2 wide receivers, 1 defensive end, 1 offensive tackle, and 1 center, over the course of 17 practice and game days. We also video analyzed the data from the high school players (n=49) over the course of 17 practice and game days. The high schoolers typically play multiple positions, unlike collegiate players, so our data sample included almost all football positions, except a kicker. The collegiate football games followed the National Collegiate Athletic Association (NCAA) Football Rules, and the practices were typically 2 hours in duration. The highschool games and practices were also approximately 2 hours in duration, in accordance with the California Interscholastic Federation (CIF). The human subject protocols were approved by the Stanford Administrative Panels for the Protection of Human Subjects (IRB-45932 and IRB-46537). We conducted data collection and performed video analysis in accordance with the institutional review boards' guidelines and regulations.

The obtained kinematics data was used to train and validate an impact classifier in part based on the method published by Wu and coworkers for detecting field football head impacts [31,32]. Retraining the classifier was important because the MiG2.0 does not have the same mechanical and electrical design as previous iterations of the device [33]. In our dataset, we included video-verified head impacts, while video-verified noncontact recordings were defined as noncontact events. To assign the mouthguard kinematics to one of the aforementioned classes, we synchronized mouthguard event timestamps to timestamps from practice and game videos, and all events (as detected by over 10 g linear acceleration in any direction) that were on camera were video verified as either a real impact or noncontact event by members of the research team. Only data recordings where contact (or noncontact) could be clearly and confidently identified were included in the training/validation dataset. Videos were acquired from at minimum 2 angles per field at 30 frames per second. However, as over roughly 70% of these potential impacts occurred out of view for the cameras, an improvement to the tracking of players would yield a larger dataset. In the future, we are hoping to track individual players throughout the entire practice or game in order to capture many more impacts and to get a better sense of the impacts that are off camera.

## 2.5 MiGNet: Neural Network Classifier for Impact Detection Validation on Field Impacts

In this section, we present a deep learning approach to automatically extract important features to distinguish between real and false impacts to a higher degree of accuracy using mouthguard time series data as the input. Deep learning has become common in many areas of biomedical/biomechanical research [34,35,36] where the signals generated are complex and a more automated feature extraction process is desired. One trade off to this approach, compared with previous work, such as the SVM classifier used in [33] is that the features are no longer selected or directly related to any engineered metric. The SVM model used in [33] is compared to this new deep learning model for our new dataset with the advantages and disadvantages compared.

### 2.5.1 The dataset of mouthguard signals

The dataset consisted of CSV files that contain the recorded time series signal of recorded mouthguard kinematic data. This included 6 variables, the x,y,z components of linear acceleration as well as the three components of angular acceleration. These signals were 200ms in length with the triggering point marking the point 50ms into the signal and 150ms before the end (see Section 2.4). In the work in [33], the SVM was trained on signals with 100ms in length so this necessitated a retraining of the model.

The dataset used to train this new SVM and the new MiGNet neural network consisted of 358 true impacts and 500 false impacts. These video-verified impact signals were obtained from the MiG2.0 mouthguards worn by the athletes. The events were classified as true and false impacts by video-confirmation, as described in previous sections. For the support vector dimensions, a feature extraction process was conducted for the data with a column selection process from [28] used to identify the most relevant features.

**2.5.2 Augmentation and Imbalanced Data Strategy for MiGNet Training**

Due to the complexity of the signals and the imbalance in the data, two strategies were executed to achieve the best performance for the network training. Firstly, to increase the number of signals used to train with, an augmentation of the signals was performed whereby the signal of a kinematic variable was shifted along the time axis by a unit of milliseconds. The total number of augmented signals from one original sample was a controlled variable. To ensure the signals were still representative of realistic impacts, and to ensure sufficient data for training, 5 augmented samples were obtained, meaning that the signal was offset by 1-5ms for each collected signal. However, it is important to note that the test set used for later evaluation was unaltered and not used in training.

In addition to the modest dataset size that was available, this dataset was also imbalanced. This resulted in suboptimal performance of the network originally. By calculating the different weighting for each class such that their importance is equally treated in the loss function used during training, this imbalance can be treated more fairly. This essentially allows for the loss-function to pay more attention to losses for the true events, and less so for the false events. More details of treating imbalanced classes can be found in [37].

As the input data was in a 1D time-varying format, the network architectures were chosen accordingly. Several variations of network layer size and architecture were iterated until the final choice of network, shown in the schematic of Fig. 3. This included a series of 1D convolution layers that were later combined into a 2D output where 2D convolutions were then performed, until culminating in a softmax output layer. 1D convolutions were used in preference to Long Short-term Memory (LSTM) due to the efficiency of 1D convolution layers [38,39]. Stochastic gradient descent was used to train the network using the Keras framework with Google's TensorFlow as the background engine. Batch sizes of 32 were iterated for 20 epochs with a learning rate of 0.01 and momentum of 0.9.

Finally, the comparative performance of the neural net and the SVM was evaluated using sensitivity, specificity, accuracy and precision as described below:

$$Sensitivity = \frac{TP}{TP + FN} \quad Specificity = \frac{TN}{TN + FP}$$

$$Accuracy = \frac{TP + TN}{TP + FP + TN + FN} \quad Precision = \frac{TP}{TP + FP}$$

True positive (TP) was an impact classified as impact, true negative (TN) was a nonimpact classified as a nonimpact, a false positive (FP) was a nonimpact classified as an impact, and a false negative (FN) was an impact classified as a nonimpact.

## 3. Results

### 3.1 MiGNet vs. SVM Impact Detection Validation

The MiGNet was tuned using a "greedy" optimization scheme for number of 1D convolutional layers, 2D convolutional layers and type of final layer. After testing 1, 2, 3, and 4 1D convolutional layers we found minimal difference in performance. Similarly, the number of 2D convolutional layers did not significantly affect performance metrics. For the final iteration of the MiGNet, two 1D convolutional layers and one 2D convolutional layer was used in order to increase the computational speed of the net by providing fewer parameters. For the last hidden layer, Global Average Pooling resulted in the best accuracy due to its emphasis on connection between features, maps, and categories. The SVM features were extracted using the inbuilt MATLAB function, *sequentialfs*, and the SVM itself was trained using the inbuilt MATLAB function, *fitcsvm*. Testing these two models on an out of sample set of 65 impacts and 100 non impacts yielded the results shown in Table 1. For this dataset, the two methods are comparable in the four metrics, with the MiGNet slightly outperforming the SVM overall. However, with more data, these machine learning models typically improve performance. Unlike the neural network, though, for the SVM a larger dataset requires that the feature selection process be repeated. As this feature selection process involves a matrix decomposition for each feature, this process is incredibly expensive and largely infeasible where other options are available. For the MiGNet, adding more data to train does still have the effect of increasing the training time, but at a much more efficient rate. In an attempt to improve the MiGNet performance, a larger dataset was applied to the training, with 3730 nonimpact and true impact signals. Retraining on this larger dataset for the SVM would take an infeasible amount of time; however, the MiGNet model was retrained within the time it took to train the smaller SVM. With this extra data, the MiGNet's final performance metrics are shown in Table 1 with the MiGNet reporting 96% accuracy, 86% precision, 99% specificity, and 76% sensitivity on the out of sample test set.

## 4. Discussion

**4.1 Shared FITBIR Platform**

The purpose of this work was to propose an open source platform for collecting, analyzing, and sharing head impact data, including head impact kinematics and the resulting health outcomes, such as sideline neurocognitive test results and whether a concussion occurred. This proposed platform would help investigators to further study links between head impact kinematics and various proposed injury metrics, thus giving further insights into the biomechanical mechanisms of concussive and subconcussive impacts. Furthermore, in this manuscript, we described a wearable device to accurately detect and capture the features of high-speed head impacts and a pipeline to easily integrate the data captured with this device into the FITBIR platform. The discussed device has been distributed to a selected group of investigators currently enrolled in an NIH funded project aimed at elucidating the biomechanical underpinnings of concussion. The standardization of a high quality concussion dataset to establish a link between head impact kinematics and brain injury is paramount especially since these injuries are rare for any one individual. Given this, it appears that the necessary data can be more quickly and effectively gathered through large-scale collaborative efforts involving multiple institutions through the dissemination of a validated head impact measurement device. The MiG2.0 collects data in a standardized form employing the NINDS recommended CDE, thus facilitating collaboration and interconnectivity with informatics platforms such as FITBIR, where data will be released in an open-source format at the time of project completion in the next couple of years. We believe that the collective knowledge gained from this collaborative effort will improve our understanding of concussion and aid in determining the now elusive link between head kinematics and brain injury.

**4.2 Evaluation of the MiGNet Impact Detection Algorithm**

Combining the use of head impact sensor data with independent video analysis, we trained and evaluated an impact detection classifier: a convolutional neural net trained and tested with high school and collegiate field data. The MiGNet architecture was developed using a greedy optimization scheme, and a parameter sweep was used to find the optimal filter size, kernel width and dropout percent.

As a deep learning product, the MiGNet benefits from a large availability of data, and while the augmentation strategy described earlier allows for sufficient training data, additional real-world data will improve this platform's performance. As the proposed integrated platform becomes widespread amongst the research community, the head impact dataset will enable more accurate training of the neural net, leading to enhanced impact detection through MiGNet workflow. Though it may take years to obtain a sizable dataset, the performance of the MiGNet gives promise that high-accuracy impact detection may be further achieved based on this methodology.

### 4.3 Limitations

The classifier described in this study was only evaluated for use with the MiG2.0 in American football. Retraining of the classifier is potentially necessary for any device that does not have the same mechanical and electrical design as the MiG2.0. We also recognize the need to test impact detection in other contact sports, both helmeted and unhelmeted to determine whether retraining the algorithm is a necessity given that other sports will likely involve forms of contact and/or player behaviors that may not have been analyzed by the current classifier. Moreover, using video information to establish ground truth data labels has some limitations that could perhaps be addressed by advances in computer vision. In this study, we rigorously checked all video labels to ensure high confidence in the training set and only used data that was confidently and clearly verified on video. Almost all mouthguard data recordings from players buried in a pile were excluded, although helmet contact is likely to occur in such situations and may be of a unique nature, which we have disregarded. Thus, we have limited our training to only very visible events. We would also like to note that we don't consider potential impacts that were missed by the mouthguard, which tend to be because they are below the minimum threshold set to trigger an impact, we instead focus only on the mouthguard recorded events for this study.

### 5. Conclusion

There are many unanswered questions underlying the biomechanics of brain trauma: we do not know what force levels could lead to a concussive event or whether this threshold changes with age, gender, or type of activity. For the research field to answer all of these questions, an integrated platform for collecting, post-processing, and sharing standardized, high-quality head impact data is essential. In previous work, we have demonstrated that the MiG2.0 can collect high-quality head impact kinematics in helmeted contact sports like American Football. In the current study, we provided a validated machine learning algorithm for impact detection and classification of on-field events. Overall, by collecting standardized common data elements, the MiG2.0 (and MiGNet) and FITBIR platform facilitate collaboration across investigators. This feature makes the MiG2.0 and FITBIR platform appropriate for enabling and fostering across-borders and multi-institutional collaboration efforts to produce a standardized, high-quality database for head impacts, including the head impact kinematics and resulting health outcomes, such as whether concussion occurred. We are hopeful that the knowledge gained from this dissemination effort will improve our basic understanding of the science of concussion.

**Acknowledgements**

The authors would like to thank Stanford Athletics and the football team for their support in this research. We would also like to thank the Camarillo Lab interns who helped in data collection.

Research reported in this publication was supported by the Stanford Center for Clinical and Translational Research and Education, the Pac-12 Conference's Student-Athlete Health and Well-Being Initiative, Taube Stanford Children's Concussion Initiative, the Office of Naval Research Young Investigator Program, the Stanford Child Health Research Institute, and the National Institute of Neurological Disorders And Stroke of the National Institutes of Health under Award Number R24NS098518. The content is solely the responsibility of the authors and does not necessarily represent the official views of the National Institutes of Health.

**Author Contributions**

All authors have made substantial contributions to (1) conception and design, acquisition of data, and/or analysis/interpretation of data and (2) drafting and/or critical revisions of the manuscript.

**Competing Interests**

The authors declare that the research was conducted in the absence of any commercial or financial relationships that could be construed as a potential conflict of interest. However, Stanford University has filed several patents in relation to the work mentioned in this manuscript.

**Corresponding Author**

Correspondence should be addressed to David B. Camarillo.

# Figures:

**FIGURE 1**: **Visual representation of the proposed integrated platform for collection, processing and sharing of mTBI data**. (a) Distribution of MiG2.0 to partner investigators who enroll in the NIH funded study (in collaboration with FITBIR) to generate an open-access platform for sharing standardized head kinematic and concussion data; (b) MiG2.0's are deployed to collect field data; (c) Raw data can be accessed and downloaded from the mouthguards directly by the investigators using the custom-made BiteMac application; (d) The raw data will be processed with our impact detection algorithm to distinguish true impacts from fake impacts; (e) Uploading to FITBIR platform can be carried out by investigators to share mTBI data and encourage other investigators to join the study as well.

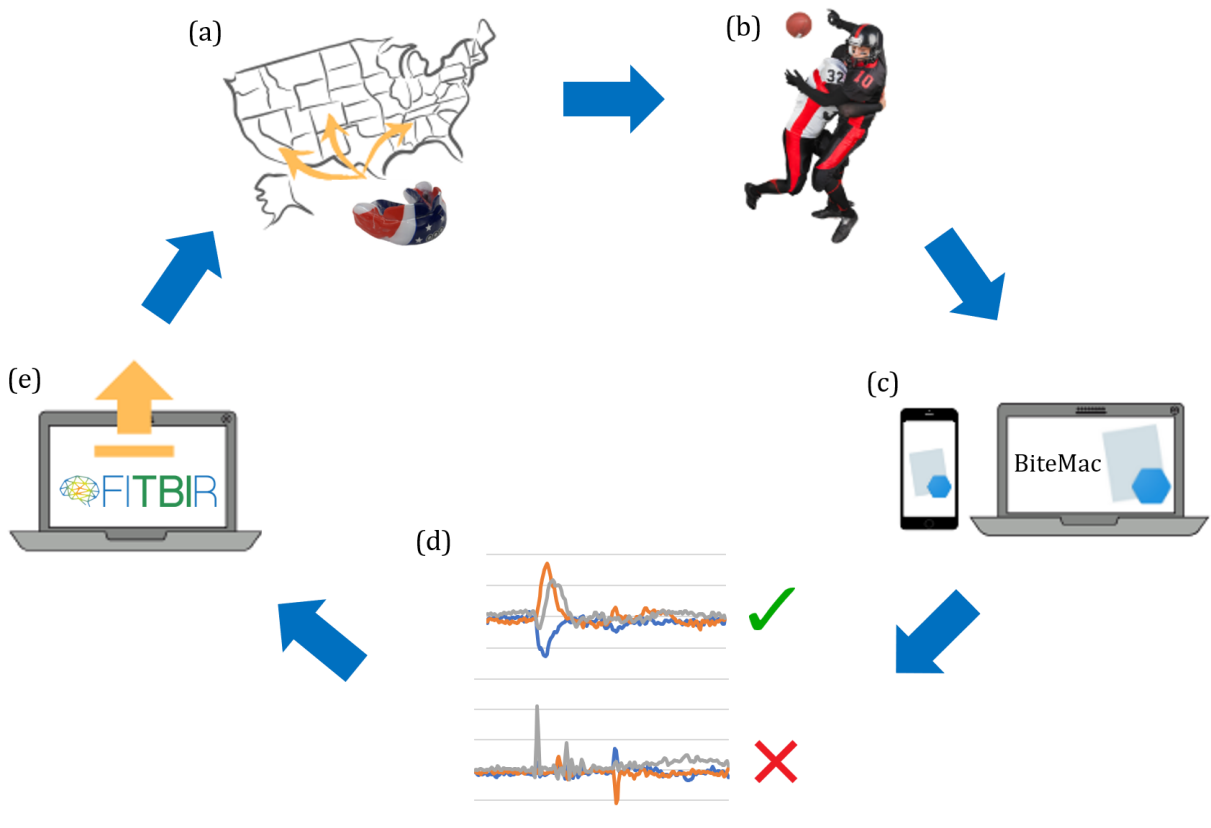

**FIGURE 2**: **MiG2.0 design.** (a) The Stanford instrumented mouthguard is a custom made mouthguard instrumented with a triaxial accelerometer and a triaxial gyroscope for measurements of head kinematics. (b) The sensory board is located at the incisors. For in mouth sensing the MiG2.0 is equipped with an infrared sensor. Unique shock absorbers are placed at the molars to reduce external disturbances.

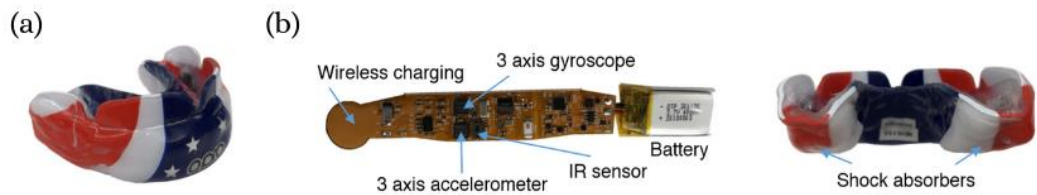

**FIGURE 3: MiGNet Architecture Schematic.** The 1D convolutional layers act to extract high-level features of the motion signal, feeding into a 2D convolution which fuses the sensor signals together.

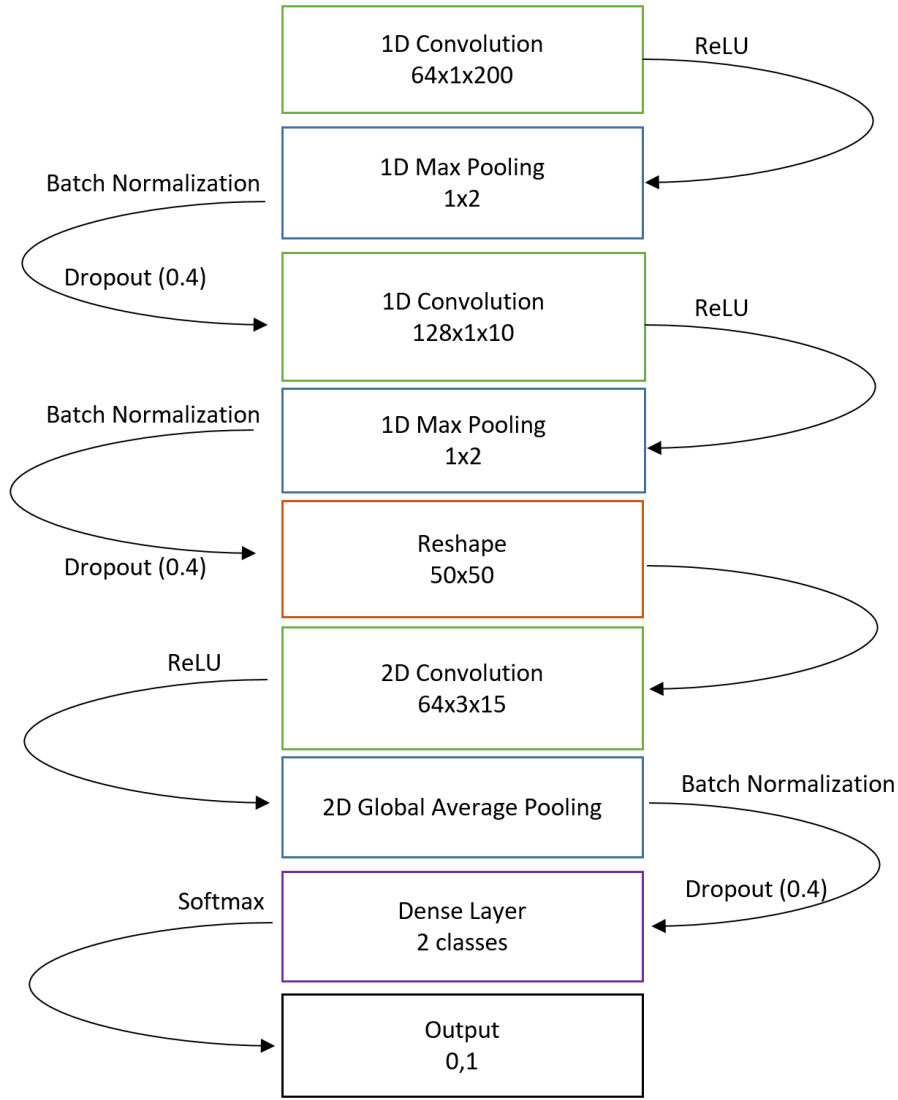

# TABLES:

**TABLE 1: MiGNet and SVM performance.** Performance measures for the trained MiGNet and SVM showing the performance on the initial dataset (Test 1) and the larger dataset used to improve the MiGNet (Test 2).

|              | Test 1 - SVM | Test 1 - MiGNet | Test 2 - MiGNet |
|--------------|--------------|-----------------|-----------------|
| Dataset size | 165          | 165             | 512             |
| Sensitivity  | 86%          | 97%             | 76%             |
| Specificity  | 94%          | 90%             | 99%             |
| Accuracy     | 91%          | 93%             | 96%             |
| Precision    | 90%          | 86%             | 86%             |